%% file: iclr2027_conference.tex
\documentclass{article} 
\usepackage{iclr2027_conference,times}

\input{math_commands.tex}

\usepackage{hyperref}       
\usepackage{url}            
\usepackage{booktabs}       
\usepackage{amsfonts}       
\usepackage{nicefrac}       
\usepackage{microtype}      
\usepackage{xcolor}         
\usepackage{multirow}
\usepackage{graphicx}
\usepackage{xspace}
\usepackage{amsmath}
\usepackage{subfig}
\newcommand{\sysname}{\textit{GA-SVD}\xspace}
\title{GA-SVD: Generation-aware Singular Value Decomposition for LLM Compression}

\author{Shengkun Tang$^1$, Cong Zeng$^2$, Zongfang Liu$^{3,4}$, Dan Alistarh$^{5}$, Zhiqiang Shen$^1$ \\
$^1$MBZUAI, $^2$Technology Innovation Institute UAE, $^3$Zhejiang Univeristy, \\$^4$Westlake University, 
$^5$ISTA\\
}

\iclrfinalcopy
\begin{document}

\maketitle
\lhead{Preprint}
\input{Sec/abs}
\input{Sec/Intro}

\input{Sec/related}
\input{Sec/method}
\input{Sec/exp}

\input{Sec/conclusion}

\section*{AI use statement}

In this work, we used generative AI tools for assist in the writing of proofs, design or provide feedback on research  methodology or experiments, implement methods.
We have not used generative AI tools for assist with translation, support qualitative and thematic data analysis, interpret results, propose or refine hypotheses,
and Generate synthetic data sets, help develop theoretical models or conceptual frameworks, formulate mathematical claims, provide critical ingredients for proving mathematical claims, clean and reformat dataset, are not applicable to this work.
Additionally, we used generative AI tools for create or edit software code. We have reviewed all AI-assisted work. We checked the LLM-generated code carefully and tested all results multiple times. We take responsibility for the final content of this work,
including text, claims or artifacts produced with the aid of generative AI.

\section*{Ethical Statement}
This work aims to improve the efficiency of LLM deployment by reducing memory and inference costs. Positive impacts include improved accessibility of capable language models in resource-constrained environments and reduced computational cost. Potential negative impacts include making capable generative models easier to deploy for misuse, such as spam or misinformation generation. The method does not introduce new model capabilities, and standard safeguards for the underlying LLMs should still be applied after compression.

\section*{Reproducibility statement}
We describe the formulation and algorithm of GA-SVD in Section~\ref{sec:method}, and provide the experimental settings and implementation details in Section~\ref{sec:exp}. These descriptions and analyses clarify the conditions under which our results are obtained and support their reproduction.

\bibliography{iclr2027_conference}
\bibliographystyle{iclr2027_conference}

\input{Sec/appendix}

\end{document}

%% file: math_commands.tex
\usepackage{amsmath,amsfonts,bm}

\def\eqref#1{equation~\ref{#1}}

\def\1{\bm{1}}

\DeclareMathAlphabet{\mathsfit}{\encodingdefault}{\sfdefault}{m}{sl}
\SetMathAlphabet{\mathsfit}{bold}{\encodingdefault}{\sfdefault}{bx}{n}



%% file: Sec/abs.tex
\begin{abstract}
Large language models (LLMs) demonstrate strong generalization across a broad range of reasoning and generation tasks, but their large parameter counts impose substantial memory and computation costs, hindering deployment in practical resource-constrained settings. One-shot low-rank compression is a promising solution since it reduces model size and inference cost effectively. However, existing SVD-based methods primarily target multiple-choice benchmarks and often fail to preserve performance on generation tasks. Moreover, their rank allocation strategies are typically heuristic. We propose \sysname, a generation-aware low-rank compression framework for LLMs. First, we introduce Generation-Aware Weight Whitening (GAWW), which leverages gradients from on-policy knowledge distillation to construct whitening transformations aligned with generative behavior. Second, we develop an evolutionary global rank allocation method that searches for effective layer-wise rank allocation under a given global compression ratio. Across extensive experiments on Mistral-7B, Qwen3 series, and Llama-3.3 models, evaluated on both multiple-choice and generative benchmarks, \sysname consistently outperforms existing SVD-based baselines and achieves state-of-the-art (SoTA) results. We further demonstrate scalability by applying our method to models with up to 70B parameters. Code will be released.
\end{abstract}

%% file: Sec/Intro.tex
\section{Introduction}

Recently, large Language Models (LLMs)~\citep{yang2025qwen3,qwen35blog, llama3modelcard}  can generate fluent, human-like content and generalize across a wide range of reasoning and generation tasks. However, the powerful performance comes from the large-scale model size, typically billions and even trillions of model parameters. The high computation and memory cost hinders the real-world application of LLMs, especially in memory-limited and latency-sensitive scenarios.

A line of work explores one-shot methods for compressing LLMs, including numerical quantization~\citep{frantar2022gptq, egiazarian2024extreme, tang2024bi}, network pruning~\citep{frantar2023sparsegpt, kurtic2023ziplm, tang2026slimqwen} and weight low-rank compression ~\cite{wang2025svdllm, wang2025svd2}. Among these techniques, low-rank compression typically applies singular value decomposition (SVD) to reduce model size and computational cost by truncating singular values and factorizing large weight matrices into products of smaller low-rank matrices. Unlike quantization and unstructured pruning, low-rank compression does not require custom inference kernels and can provide acceleration across standard hardware platforms. Moreover, low-rank compression is orthogonal to other compression techniques, such as quantization, and can therefore be combined with them for more aggressive model compression.

A series of SVD-based compression methods obtain satisfactory performance while reducing the model computation cost. For example, SVD-LLM~\citep{wang2025svdllm} connects singular value with compression loss by input data whitening, which decomposes the input-whitened weights and truncates the weights with the identical ranks. SVD-LLM v2~\citep{wang2025svd2} further improves the performance by analyzing the sensitivity of each layer and allocating the different compression ratios for each weight. However, there are still critical limitations of existing methods.  First of all, Existing work primarily evaluates compressed models on multiple-choice question (MCQ) benchmarks, such as HellaSwag~\citep{zellers2019hellaswag} and MMLU~\citep{hendrycks2020measuring}, while paying less attention to generation-oriented tasks, such as GSM8K~\citep{cobbe2021training} and MBPP~\citep{austin2021program}. We observe a significant performance degradation on generation benchmarks, as shown in Figure \ref{Fig_method:finetune}. We provide additional results on open-ended summarization using CNN/DailyMail in the Appendix~\ref{app:more_exp_res}. Moreover, current adaptive rank allocation methods~\citep{li2025adasvd,gao-etal-2024-adaptive} are based on heuristics, which may be suboptimal under a global compression budget.

\input{Sec/fig_sec/motivation}
To address the limitation of existing works, we propose \sysname, a generation-aware singular value decomposition framework, inspired by on-policy knowledge distillation~\citep{agarwal2024policy}. Technically, we introduce \emph{Generation-Aware Weight Whitening} (GAWW), which guides the weight matrix whitening by the gradient of on-policy KD.  The central idea is to make the whitening geometry reflect the model's generative behavior, rather
than relying only on offline calibration statistics.
To achieve this, we estimate layerwise whitening statistics from trajectories generated by the compressed
student model itself.
Unlike prior input-whitened SVD methods, our method introduces an additional output-side
sensitivity covariance, which measures how strongly perturbations in a layer output affect a generation-aware objective.
 These two statistics define a bi-sided compression metric, and under this metric the compressed layer is obtained by applying truncated SVD to a transformed weight matrix. Moreover, to obtain effective rank allocation under a global compression budget, we formulate rank selection as a search problem and introduce evolutionary global rank allocation via evolutionary search~\citep{tang2025darwinlm, liu2026evoesap}. Given a global compression ratio, we randomly mutate rank allocations to generate a set of offspring candidates at each generation using predefined mutation steps. Based on the fitness evaluation, the best candidate is selected as the final rank allocation.

Empirically, we validate our method on a wide range of models, including Mistral, Qwen, and Llama series models. Our compressed models are evaluated on both MCQ and generative benchmarks. Compared with existing methods, our method achieves the SoTA performance on both MCQ and generative benchmarks. Specifically, on Mistral-7B with 20\% compression, GA-SVD achieves the best coding average, GSM8K accuracy, MCQ average, and perplexity among compressed models. Compared with SVD-LLM, GA-SVD outperforms all existing baseline methods on both MCQ and generative benchmarks. The gains remain consistent under more aggressive 40\% compression, demonstrating strong robustness. Across Qwen3 models from 8B to 32B, GA-SVD provides substantial improvements, especially on generation-oriented benchmarks. Moreover, our method is highly scalable and we further scale our method on compressing larger models up to 70B parameters. We summarize our contributions as follows:

\begin{itemize}
    \item We propose \sysname, a generation-aware low-rank compression framework for LLMs. In particular, we introduce Generation-Aware Weight Whitening (GAWW), which incorporates on-policy knowledge distillation signals into the whitening process, enabling compressed models to better preserve generation capabilities.

    \item We introduce an evolutionary global rank allocation method for low-rank compression. Instead of relying on heuristic layer-wise allocation rules, our method formulates rank allocation as a global search problem and discovers effective rank configurations under a target compression budget.

    \item Extensive experiments across the Mistral-7B, Qwen3, and Llama-3.3 families, evaluated on both multiple-choice and generative benchmarks, demonstrate that \sysname\ achieves state-of-the-art performance among existing SVD-based LLM compression methods, while also scaling to models with up to 70B parameters.
\end{itemize}

%% file: Sec/fig_sec/motivation.tex
\begin{figure*}[t]
\begin{center}
\subfloat[Mistral-7B]{\includegraphics[width=0.33\textwidth]{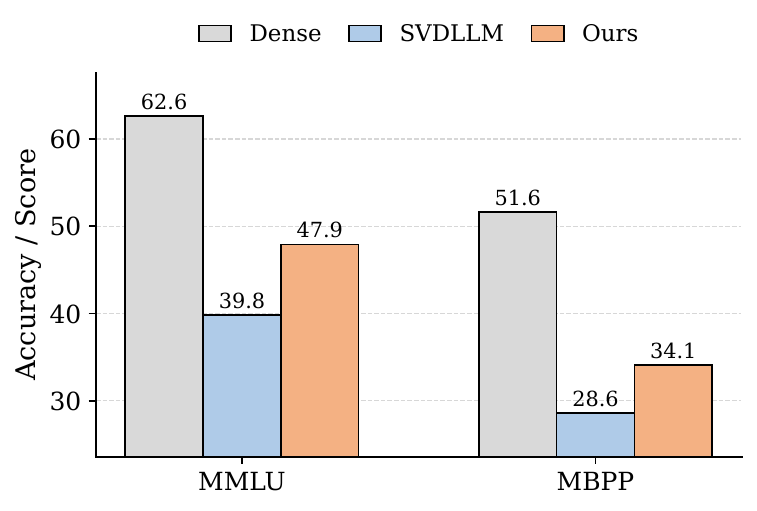}}
\subfloat[Qwen3-8B]{\includegraphics[width=0.33\textwidth]{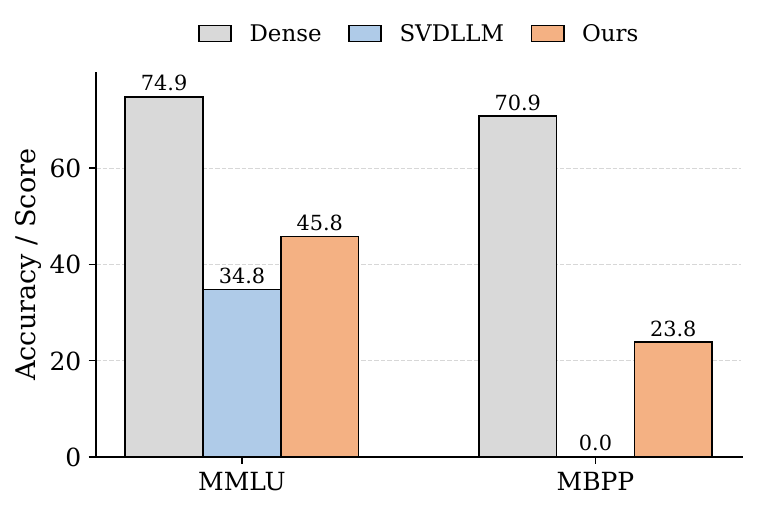}}
\subfloat[Qwen3-14B]{\includegraphics[width=0.33\textwidth]{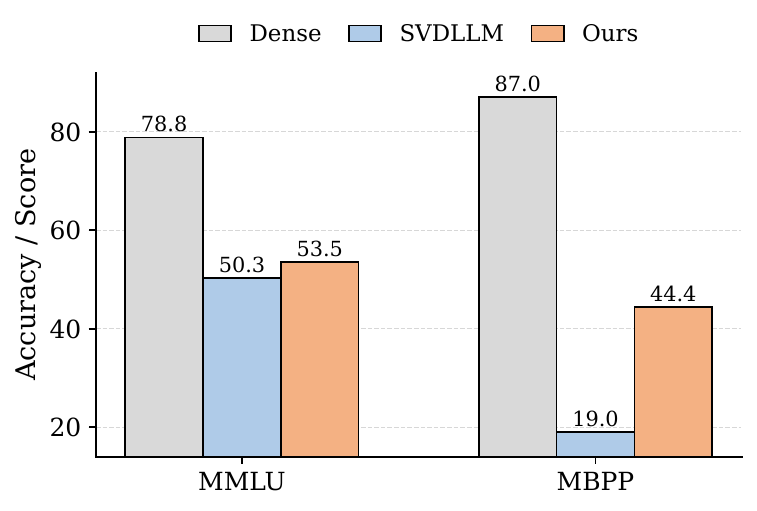}}\\ 
\end{center}
\vspace{-1em}
\caption{Existing SVD-based compression degrades generation performance.
Compared with MCQ benchmarks such as MMLU, SVDLLM suffers more severe degradation on generation-oriented benchmarks such as MBPP. GA-SVD substantially mitigates this degradation across different models. }
\label{Fig_method:finetune}
\vspace{-1.5em}

\end{figure*}

%% file: Sec/related.tex
\section{Related Work}

\noindent \textbf{Post-Training SVD-Based LLM Compression.}
Early work~\citep{hsu2022language} on language models introduced Fisher-weighted SVD (FWSVD), which weights singular value decomposition by parameter importance to preserve task-relevant components and achieves higher accuracy than vanilla SVD at aggressive compression. A closely related method, Activation-aware SVD (ASVD)~\citep{yuan2023asvd}, further adjusts weight matrices based on input activation statistics and layer sensitivity. ASVD applies a per-channel scaling to absorb activation outliers and then assigns different ranks per layer according to sensitivity. SVD-LLM \citep{wang2025svdllm} identifies that naive SVD causes high perplexity and instead uses a truncation-aware whitening preprocessing and sequential low-rank updates. SVD-LLM V2 \citep{wang2025svd2} further optimizes singular value truncation by assigning heterogeneous compression ratios according to theoretical truncation loss.
SAES-SVD~\citep{hu2026saes} proposes Cumulative error-aware layer compression to jointly consider local reconstruction and accumulated error compensation, and adaptive collaborative error suppression to automatically adjust the compensation weight. AA-SVD \citep{sinha2026aa} compresses LLMs by anchoring each compressed layer to the original model outputs while adapting to input distribution shifts caused by upstream compression,
These methods are attractive due to their simplicity and lack of retraining, but they often rely on uniform or heuristically determined ranks, which may be suboptimal. In practice, one must often tune ranks to avoid large performance drops. Moreover, the generation ability is ignored during compression. 

\noindent \textbf{Rank Selection Methods.} Rather than fixing ranks by rule, recent works~\citep{ji2024adaptive_bayes} learn the optimal truncation. For example, Adaptive Rank Selection~\citep{gao-etal-2024-adaptive} treats truncation as a differentiable process.
AdaSVD~\citep{li2025adasvd} adaptively compensates truncation errors and assigns layer-specific compression ratios based on layer importance.
D-Rank is proposed by~\citep{mi2025layer}, which is an SVD-based compression method that assigns different ranks to different layers according to their information density.
LLRC~\citep{sundrani2025low}, which uses a small calibration set and trainable binary masks over singular values to select ranks in a fine-tuning-free way. Similarly, Dobi-SVD~\citep{wang2025dobi} formulates SVD truncation as a differentiable step with analytically optimal update rules, enabling gradient-based refinement of the decomposition. However, these works generally require extra training cost for rank selection.

%% file: Sec/method.tex
\section{Method}
\label{sec:method}
\input{Sec/fig_sec/main_figure}
\noindent \textbf{Overview.} Our method starts from the student warm-start initialization, which is required by the on-policy distillation. After that, given a calibration dataset, we collect the input covariance and generation-aware sensitivity covariance, which are applied in the weight bi-whitening. We further utilize evolutionary search to find the optimal rank allocation setups. The overall pipeline can be found in Figure~\ref{fig:main_fig}.

\noindent \textbf{Notation.}
Let $M_T$ denote the original teacher model, and let $M_S$ denote the compressed student model.
We use $\mathcal{L}$ to denote the set of linear layers to compress.
For each layer $l \in \mathcal{L}$, the original full-precision weight is
$
W_l^\star \in \mathbb{R}^{d_{\mathrm{out},l} \times d_{\mathrm{in},l}}.
$
For token position $t$, we denote by $x_{l,t}$ the input activation of layer $l$, and by
$h_{l,t} = W_l^\star x_{l,t}$ the corresponding layer output.
We use $\mathcal{D}_{\mathrm{warm}}$ for the warm-start calibration set and
$\mathcal{D}_{\mathrm{prompt}}$ for the prompt set used in generation-aware statistics collection.
\subsection{Generation-Aware Weight Whitening for SVD Compression}

\noindent \textbf{Warm-start Initialization.}
Generation-aware whitening requires rollout statistics collected from a compressed student.
Since such statistics are unavailable before compression, we first construct an initial compressed model
using a standard input-whitened SVD procedure, following SVD-LLM~\citep{wang2025svdllm}.
Concretely, for each layer we collect activations from $\mathcal{D}_{\mathrm{warm}}$ and estimate an
input second-order matrix.
We then transform the original weight by the input-side factor, perform truncated SVD on the transformed
matrix, and finally absorb the inverse transformation back into the compressed factors.
This yields an initial low-rank approximation for every layer and defines the warm-start student
$M_S^{(0)}$.
Intuitively, the warm-start stage serves two purposes.
First, it provides a stable compressed model from which on-policy generation can be performed.
Second, it initializes the model with a standard whitening geometry before generation-aware refinement is
introduced.

\noindent \textbf{On-policy Generation-aware Whitening.}
Starting from the warm-start student $M_S^{(0)}$, we collect generation-aware whitening statistics from
trajectories sampled from the current compressed student model.
Given prompts $p \sim \mathcal{D}_{\mathrm{prompt}}$, we decode continuations using the student policy:
\begin{equation}
y \sim \pi_{M_S^{(0)}}(\cdot \mid p),
\qquad
z = [p; y].
\label{eq:gaww_rollout}
\end{equation}
Let $\mathcal{G}(z)$ denote the generated token positions in $z$, excluding the prompt tokens.
All statistics are computed only on $t \in \mathcal{G}(z)$ so that the resulting whitening geometry is aligned
with the student's own generative regime, rather than with teacher-forced text alone.

To measure which layer-output directions are important for generation, we run the teacher and the student
on the same realized sequence $z$ and compute a token-level distillation objective on generated tokens only.
In this work, we instantiate the generation-aware signal using reverse KL divergence and obtain the
corresponding gradient with respect to the layer output:
\begin{equation}
L_{\mathrm{gen}}
=
\frac{\tau^2}{N}
\sum_{z}
\sum_{t \in \mathcal{G}(z)}
\mathrm{KL}\!\left(
p_S^\tau(\cdot \mid z_{<t})
\,\|\, 
p_T^\tau(\cdot \mid z_{<t})
\right),
\qquad
g_{l,t}=\frac{\partial L_{\mathrm{gen}}}{\partial h_{l,t}}
\in \mathbb{R}^{d_{\mathrm{out},l}},
\label{eq:gaww_loss}
\end{equation}
where $p_S^\tau$ and $p_T^\tau$ are the temperature-scaled student and teacher token distributions, and
$
N = \sum_z |\mathcal{G}(z)|
$
is the total number of generated-token positions.
The role of this objective is not to update model parameters directly, but to induce a generation-aware
sensitivity signal from the student's own rollouts.
Using the generated-token activations and the corresponding output gradients, we estimate two layerwise
second-order statistics:
\begin{equation}
\Sigma_{x,l}
=
\frac{1}{N_l} \sum_{t=1}^{N_l} x_{l,t}x_{l,t}^{\top},
\qquad
\Sigma_{s,l}
=
\frac{1}{N_l} \sum_{t=1}^{N_l} g_{l,t} g_{l,t}^{\top},
\label{eq:gaww_cov}
\end{equation}
where $N_l$ is the number of generated-token samples accumulated for layer $l$.
Here, $\Sigma_{x,l}$ captures the geometry of the input space, while
$\Sigma_{s,l}$ captures the generation-aware sensitivity of the output space.

\noindent \textbf{Generation-aware Bi-whitened Low-rank Compression.}
We now describe how the generation-aware whitening statistics define the layerwise compression objective and
lead to the final low-rank approximation.
For each layer $l$, let $\widehat W_l$ denote a rank-constrained approximation of the original weight matrix
$W_l^\star$, and define the compression error and the corresponding perturbation at the layer output of token
position $t$ as
\[
\Delta W_l = W_l^\star - \widehat W_l,
\qquad
\Delta h_{l,t} = \Delta W_l x_{l,t}.
\]
Under a local empirical Fisher / Gauss--Newton approximation induced by the generation-aware objective, the
layerwise increase in loss can be approximated by the following bi-sided quadratic form:
\begin{equation}
\Delta \mathcal{L}_l(\widehat W_l)
\approx
\frac{1}{2}
\mathrm{Tr}\!\left(
\Sigma_{s,l}
\Delta W_l
\Sigma_{x,l}
\Delta W_l^{\top}
\right),
\label{eq:gaww_quad}
\end{equation}
where $\Sigma_{x,l}$ and $\Sigma_{s,l}$ denote the input covariance and sensitivity covariance,
respectively.
This objective shows that the compression error should be measured not in the Euclidean geometry of the raw
weight matrix, but in a geometry jointly induced by the input activations and the generation-aware output
sensitivities.
Accordingly, compressing layer $l$ amounts to solving the weighted rank-constrained problem
\begin{equation}
\widehat W_l
=
\arg\min_{\mathrm{rank}(A)\le r_l}
\mathrm{Tr}\!\left(
\Sigma_{s,l}
\left(W_l^\star-A\right)
\Sigma_{x,l}
\left(W_l^\star-A\right)^{\top}
\right).
\label{eq:gaww_rank_problem}
\end{equation}
To solve Eq.~\eqref{eq:gaww_rank_problem} efficiently, we first regularize the two whitening statistics and
compute their square-root factors:
\begin{equation}
\overline{\Sigma}_{x,l} = \Sigma_{x,l} + \epsilon_x I = R_{x,l}R_{x,l}^{\top},
\qquad
\overline{\Sigma}_{s,l} = \Sigma_{s,l} + \epsilon_s I = R_{s,l}R_{s,l}^{\top},
\label{eq:gaww_reg_factors}
\end{equation}
where the square-root factors are obtained by Cholesky decomposition in practice.
Using these factors, the original weighted objective can be rewritten as an ordinary Frobenius-norm low-rank
approximation problem in a transformed space.
This motivates the bi-whitened transformed weight
\begin{equation}
\widetilde W_l
=
R_{s,l}^{\top} W_l^\star R_{x,l}.
\label{eq:gaww_transform}
\end{equation}
The detailed derivation of this equivalence is provided in Appendix ~\ref{app:gaww_derivation}. After obtaining the transformed matrix, we perform singular value decomposition:
$
\widetilde W_l = U_l \Lambda_l V_l^{\top},
$
truncate it to rank $r_l$, and denote the truncated factors by
$U_{l,r}$, $\Lambda_{l,r}$, and $V_{l,r}$.
Since the transformed problem is now a standard rank-constrained Frobenius approximation, the truncated SVD
provides the optimal rank-$r_l$ solution in the transformed space.
We then map this solution back to the original parameterization by re-coloring:
$
\widehat W_l
=
R_{s,l}^{-\top}
U_{l,r} \Lambda_{l,r} V_{l,r}^{\top}
R_{x,l}^{-1}.
\label{eq:gaww_reconstruct}
$
For efficient storage and inference, the compressed layer is kept in factorized form
$
\widehat W_l = W_{u,l}W_{v,l},
\label{eq:gaww_factorized}
$
where
$
W_{u,l}
=
R_{s,l}^{-\top}
U_{l,r} \Lambda_{l,r}^{1/2},
W_{v,l}
=
\Lambda_{l,r}^{1/2} V_{l,r}^{\top} R_{x,l}^{-1}.
\label{eq:gaww_factor}
$
In this way, GAWW extends input-whitened SVD compression to a generation-aware bi-whitened setting, where
the low-rank approximation is jointly guided by the input geometry and the output sensitivity induced by
generation.

\subsection{Evolutionary Global Rank Allocation}

\noindent \textbf{Problem formulation.}
Consider a pretrained transformer with a set of compressible linear weights
$\mathcal{W}=\{1,2,\dots,M\}.$
Our goal is to determine a non-uniform rank allocation and corresponding singular-value selection for all
weights so as to minimize an end-to-end fitness under a global compression budget:
\begin{align}
\label{eq:evo_global_obj}
\min_{\{K\}} F(\{K\})
\qquad
\text{s.t.}
\qquad
\sum_{K\in \mathcal{K}} C(|K|) \le B,
\end{align}
where \(K\) denotes the retained singular-index set of a weight matrix, \(C(\cdot)\) is the parameter
cost of the compressed representation, \(F(\cdot)\) is the fitness function and \(B\) is the global budget.
For a rank-\(r\) factorization of a weight matrix \(W\in\mathbb{R}^{d_{\mathrm{out}}\times d_{\mathrm{in}}}\), the compressed parameter cost is
$
\label{eq:evo_cost}
C(r)=r\bigl(d_{\mathrm{in}}^{}+d_{\mathrm{out}}^{}\bigr).
$
Given a target compression ratio \(\rho\), we set
$
\label{eq:evo_budget}
B=(1-\rho)\sum_{W\in\mathcal{W}} d_{\mathrm{out}}^{} d_{\mathrm{in}}^{}.
$
To ensure that low-rank factorization remains meaningful for compression, we restrict the admissible ranks to
$
\label{eq:evo_rmax}
r^{\max}
=
\min(
R,\;
\lfloor
\frac{d_{\mathrm{out}}^{}d_{\mathrm{in}}^{}}
{d_{\mathrm{out}}^{}+d_{\mathrm{in}}^{}}
\rfloor),
$
where \(R=\mathrm{rank}(W)\).

\noindent \textbf{Search Space.}
A direct search over all subsets \(K\subseteq\{1,\dots,R\}\) is combinatorial.
To make the search tractable, we restrict the candidate set along two axes. First, for each weight, we define a discrete set of admissible rank levels
\begin{align}
\label{eq:evo_rank_levels}
\mathcal{R} = \{r^{(0)}, r^{(1)}, \dots, r^{(L)}\},
\qquad
r^{(\ell)} \le r^{\max},
\end{align}
where the rank levels are chosen as multiples of a step size, clipped by
\(r^{\max}\). Second, given a retained rank \(r\), we allow non-monotonic singular selection only near the truncation
boundary.
Specifically, we define a boundary window
\begin{align}
\label{eq:evo_boundary}
\mathcal{B}(r)
=
\left\{
i \,\middle|\,
\max(1,r-w+1)\le i \le \min(R,r+w)
\right\},
\end{align}
where \(w\) is a small window size.
For each weight, the retained set is parameterized as
$
\label{eq:evo_subset}
K=\{1,2,\dots,p\}\cup B', \,
B' \subseteq \mathcal{B}(r),
$
subject to the cardinality constraint
$
|K|=r.
$
Thus, the search captures two forms of non-uniformity: inter-weight non-uniformity through different
rank choices, and limited intra-weight non-uniformity through boundary selection.

\noindent \textbf{Initialization.}
Each offspring candidate compressed model is encoded as 
$
\label{eq:evo_genome}
\mathcal{O}=\{o\},\,
o=(r,z),
$
where \(r \in \mathcal{R}\) is the retained rank level and \(z\) encodes the boundary selection mask
over \(\mathcal{B}(r)\).
Decoding \(\mathcal{O}\) yields the retained singular-index sets \(\{K\}\). We initialize the search from a feasible top-\(k\) candidate.
Concretely, we first choose an initial rank allocation \(\{r^{(0)}\}\) satisfying the global
budget, and then initialize each weight with standard top-\(r^{(0)}\) truncation:
$
\label{eq:evo_init}
K^{(0)}=\{1,2,\dots,r^{(0)}\}.
$
This gives a valid starting point that already respects the bi-whitened singular ordering.

\noindent \textbf{Mutation.}
We adopt a \((1+\lambda)\) evolutionary strategy.
From the current parent genome, we generate \(\lambda\) offspring via random mutation.
Since we only consider weight-level mutation, all compressible weights belong to a single mutation pool.

\emph{Rank mutation.}
We randomly choose two weights, decrease the rank of one weight by one level, and increase the
rank of the other weight by one level:
\begin{align}
\label{eq:evo_rank_transfer}
r_i \leftarrow r_i - \delta,
\qquad
r_j \leftarrow r_j + \delta,
\end{align}
where the updated ranks must remain within their admissible rank sets and $\delta$ refers to the mutation step.
This mutation preserves the global budget approximately while enabling non-uniform redistribution of rank
across weights.

\emph{Boundary-swap mutation.}
For a chosen weight, we modify the retained subset within the boundary window.
Specifically, we swap one retained boundary index with one dropped boundary index:
\begin{align}
\label{eq:evo_boundary_swap}
K
\leftarrow
\bigl(K \setminus \{i\}\bigr)\cup\{j\},
\qquad
i \in K \cap \mathcal{B}(r),
\quad
j \in \mathcal{B}(r)\setminus K.
\end{align}
This mutation allows limited non-top-\(k\) adjustment without exploding the search space. For the offspring selection in each generation, we select the best candidate of all offspring and consider the candidate as the parent of next generation.

%% file: Sec/fig_sec/main_figure.tex
\begin{figure*}[t]
    \centering
    \includegraphics[width=0.9\linewidth]{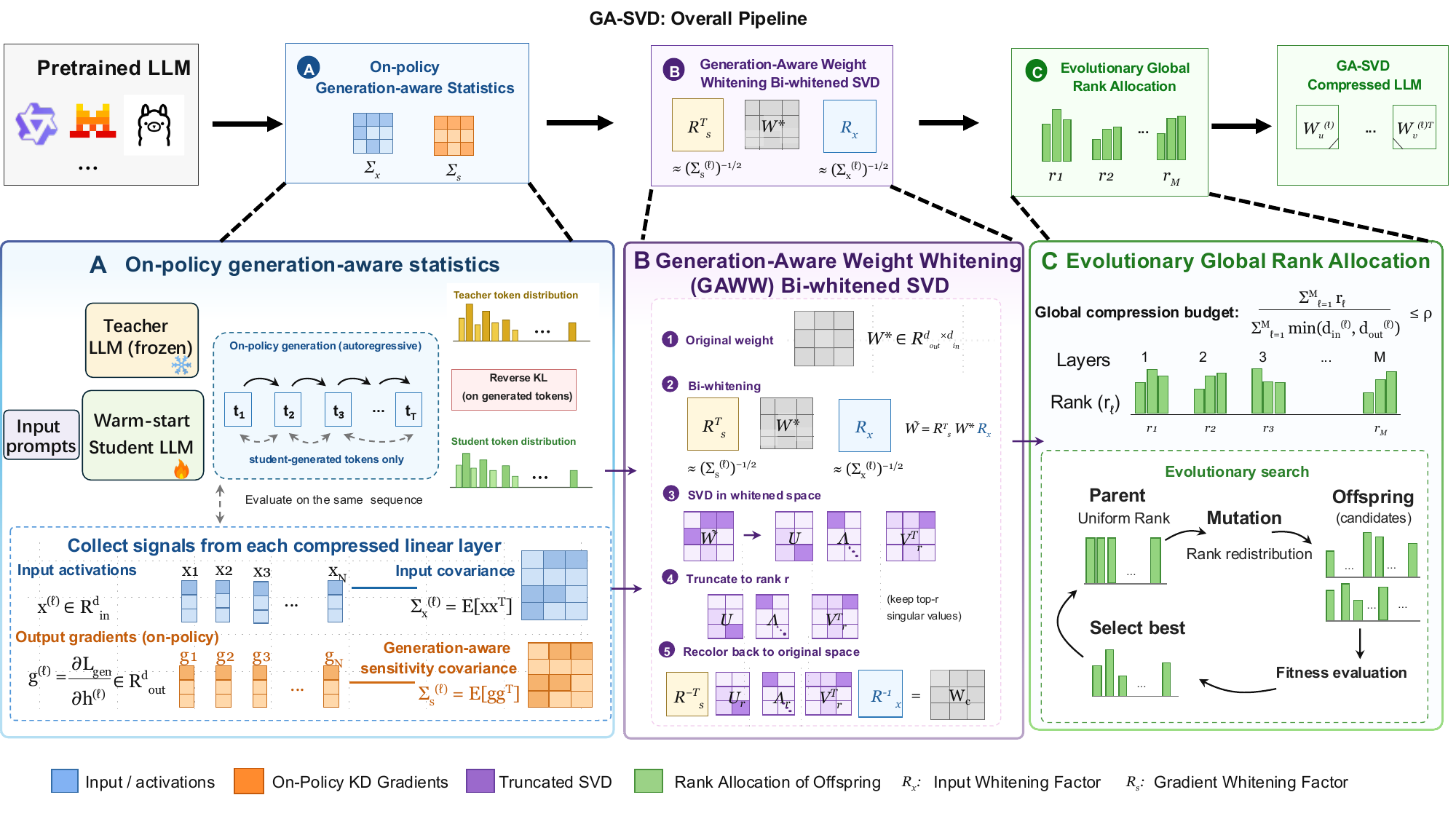}
    \vspace{-1em}
    \caption{Visual illustration of the \sysname pipeline. \sysname first collects on-policy generation-aware statistics from a warm-start compressed student using teacher-student distillation signals. It then performs Generation-Aware Weight Whitening (GAWW), where input activations and output gradients define a bi-whitened SVD objective for each compressed linear layer. Finally, Evolutionary Global Rank Allocation searches for an effective layer-wise rank configuration under a global compression budget, producing the final \sysname compressed LLM.}\
    \label{fig:main_fig}
\vspace{-2em}
\end{figure*}

%% file: Sec/exp.tex
\section{Experiments}
\label{sec:exp}
\noindent \textbf{Model \& Dataset \& Baselines.} We conduct our methods on a range of different LLMs. We compare our method with baselines on Mistral 7B~\citep{jiang2023mistral} and Qwen3~\citep{yang2025qwen3} series, including 8B, 14B and 32B models. To further show the scalability and validate our method on instruction-tuned model, we test our method on Llama-3.3-70B-Instruct~\citep{llama3modelcard} models. For the calibration dataset, we use wikitext~\citep{merity2016pointer} training set, Evol-codealpaca-v1~\citep{luo2023wizardcoder} and tulu-3-sft-personas-math~\citep{lambert2024tulu} to assess both general and domain knowledge calibration.
We compare \sysname with various baselines including ASVD~\citep{yuan2023asvd}, SVD-LLM~\citep{wang2025svdllm}, SVD-LLM v2~\citep{wang2025svd2}, Dobi-SVD~\citep{wang2025dobi} and AA-SVD~\citep{sinha2026aa}. As the closest work, we compare with SVD-LLM on each following analysis experiment. We apply the official implementation of each method with the same calibration set.

\noindent \textbf{Evaluation.}
We report the average perplexity of compressed model on Wikitext~\citep{merity2016pointer}, C4~\citep{allenai_c4} and PTB~\citep{marcus-etal-1993-building} test set. The detail results of each dataset can be found in Appendix Table~\ref{tab:more_main_table}. Moreover, compressed LLMs are evaluated with both multiple-choice questions (MCQ) and generative benchmarks including coding and math generations. For MCQ, we evaluate the compressed model on 0-shot PIQA~\citep{bisk2020piqa}, 25-shot ARC-C~\citep{allenai:arc}, 10-shot HS~\citep{zellers2019hellaswag} and 5-shot MMLU~\citep{hendrycks2020measuring} while HumanEval~\citep{chen2021evaluating} and MBPP~\citep{austin2021program} are used to evaluate the coding generation, and 8-shot GSM8K~\citep{cobbe2021training} are utilized for math generation evaluation. HumanEval and MBPP are evaluated by \textit{evalplus}~\citep{evalperf,evalplus} packages and others are executed by \textit{lm-evaluation-harness} packages.

\noindent \textbf{Implementation Details.} For calibration, we use 1024 samples with the sequence length of 2048 from wikitext, Evol-codealpaca-v1 and tulu-3-sft-personas-math. For large models, Qwen3-32B and Llama-3.3-70B-Instruct, we apply 256 samples for faster calibration. For on-policy generation-aware whitening, we use 256 samples with the prompt length of 128 and rollout length 512. The temperature is 1.0. We compress the model with 0.2 and 0.4 compression ratio. For evolutionary rank allocation, we apply 32 samples for fitness computation. We utilize KL divergence as the fitness function.  For each offspring generation, 16 candidates are generated with 10 times mutations randomly. The rank mutation step is 128 and the boundary window  \(w\) is 32 for faster searching. We search the rank allocation setup for 100 generations. All experiments, including main results and analysis are conducted on a  4$\times$H100 machine.

\subsection{Main Results}
\input{tables/main_table}
The main results are shown in Table~\ref{tab:main_table}. We compare GA-SVD with representative SVD-based compression baselines on Mistral-7B at a 20\% compression ratio. GA-SVD achieves the best performance across all major evaluation categories, including coding generation, mathematical reasoning, MCQ benchmarks, and perplexity. On coding tasks, GA-SVD improves the code average to 27.4, outperforming SVD-LLM, SVD-LLM v2, Dobi-SVD, and AA-SVD. The gains are consistent on both HumanEval and MBPP, suggesting that GA-SVD better preserves generation-oriented capabilities after compression. For mathematical reasoning, GA-SVD obtains 18.5 on GSM8K, exceeding all baselines. On MCQ benchmarks, GA-SVD achieves the highest average score of 51.7, with clear improvements on PIQA, MMLU, ARC-C, and HellaSwag. In particular, compared with SVD-LLM, GA-SVD improves MMLU from 39.8 to 47.9 and HellaSwag from 44.3 to 53.0, showing that the proposed method also benefits standard multiple-choice evaluation. GA-SVD further achieves the lowest perplexity among compressed models, reducing PPL from 50.4 with SVD-LLM to 37.8. Overall, these results demonstrate that GA-SVD preserves both generative and discriminative abilities more effectively than prior SVD-based methods.

\noindent \textbf{Performance on Different LLMs Series and Sizes.}
 \input{tables/main_larger}
We report the results on different model series and sizes with 20\% compression ratio, as shown in Table~\ref{tab:large_scale}. The results show that GA-SVD consistently outperforms SVD-LLM across different model families and sizes, including Qwen3-8B, Qwen3-14B, Qwen3-32B, and Llama-3.3-70B-Instruct. The improvements are observed on coding, math, MCQ benchmarks, and perplexity, indicating that GA-SVD better preserves both generative and discriminative capabilities after compression. Notably, the gains are especially large on generation-oriented tasks such as HumanEval, MBPP, and GSM8K, which supports the effectiveness of generation-aware whitening. This is important because generation benchmarks require the compressed model to maintain coherent autoregressive behavior over long output sequences, making them more sensitive to compression-induced distribution shifts than standard MCQ tasks. Across the Qwen3 series, GA-SVD remains effective as the model size increases from 8B to 32B. Note that both SVD-LLM and GA-SVD exhibit substantial performance degradation on Qwen3-8B, particularly on coding and math tasks. This reflects lower parameter redundancy than in larger models, making rank reduction more disruptive to autoregressive reasoning. We provide further analysis in Appendix ~\ref{app:more_exp_res}. The consistent reduction in perplexity further shows that GA-SVD better maintains the language modeling distribution after low-rank approximation. Beyond base models, GA-SVD also scales successfully to the 70B instruction-tuned Llama-3.3 model, improving downstream performance and reducing perplexity compared with SVD-LLM.

\noindent \textbf{Performance on Different Compression Ratio.}
\input{tables/table_ratio} We provide the results on different compression ratios, as shown in Table ~\ref{tab:compression_ratio}. The results show that GA-SVD consistently outperforms SVD-LLM at both 20\% and 40\% compression ratios. At 20\% compression, GA-SVD improves the coding average from 21.9 to 27.4, GSM8K from 14.9 to 18.5, MCQ average from 45.4 to 51.7, and reduces perplexity from 50.4 to 37.8. Under the more aggressive 40\% compression setting, GA-SVD still achieves consistent gains, improving the coding average from 9.8 to 13.1, MCQ average from 34.9 to 37.0, and reducing perplexity from 137.4 to 97.5. These results indicate that GA-SVD is more robust to increasing compression ratio and better preserves both generative and MCQ capabilities than SVD-LLM. We also note that although GA-SVD substantially improves over SVD baselines, a non-trivial gap to the dense model remains, especially for smaller models and aggressive compression.

\subsection{Analysis}

\noindent \textbf{Rank Allocation Distribution.}
\input{Sec/fig_sec/mistral_rank_allocation}
We visualize the layer-wise ranks assigned by the evolutionary global rank allocation strategy for representative weight matrices, as shown in Figure~\ref{Fig_exp:rank_alloc}. The searched allocations are highly non-uniform across layers, showing that different layers and projection types have distinct sensitivity to low-rank compression. In particular, Down\_Proj, K\_Proj, and O\_Proj exhibit different rank ranges and variation patterns, suggesting that a fixed or heuristic rank allocation cannot fully capture the heterogeneous redundancy of transformer weights. Moreover, these observations indicate that rank sensitivity is heterogeneous along both dimensions: it varies not only across transformer layers, but also across different weight matrices within the same layer, further motivating weight-wise global rank allocation rather than layer-only or uniform allocation strategies. The evolutionary search automatically redistributes the rank budget toward more sensitive layers while assigning lower ranks to more compressible ones, leading to a better global trade-off under the target compression constraint.  More rank allocation distribution of different models and weights can be found in the Appendix~\ref{sec:more_vis}.

\noindent \textbf{Runtime and Speed Analysis.}
\input{tables/speedup} We report the efficiency of GA-SVD on a single H100 GPU with batch size 8192, as shown in Table~\ref{tab:speedup}. Compared with the dense Mistral model, GA-SVD at 40\% compression ratio increases inference throughput from 22,931 to 27,720 tokens/s, yielding a 20.8\% speedup, while reducing GPU memory usage from 13.6 GB to 8.3 GB, corresponding to a 38.9\% reduction. In addition, GA-SVD completes compression in 5.2 hours, which is substantially faster than Dobi-SVD’s 112.4 hours. The time is the summation of data whitening, SVD compression and evolutionary search.  These results show that GA-SVD not only improves model compactness and inference efficiency, but also keeps the compression procedure computationally practical.

\subsection{Ablation Study}
We provide the ablation study in Table~\ref{tab:ablation}. The results verify the contribution of each component in GA-SVD. Removing GAWW decreases performance on all generative benchmarks, indicating that generation-aware whitening helps preserve generation ability during compression. This suggests that relying only on input-side statistics is insufficient for generation-oriented tasks, since it does not capture how layer-output perturbations affect the model’s generated responses. Removing ERA causes a larger drop on coding tasks, reducing HumanEval from 20.7 to 16.7 and MBPP from 34.1 to 30.9, which shows the importance of global rank allocation beyond heuristic rank assignment. This also indicates that different weights and layers have heterogeneous compression sensitivity, and that automatically redistributing the rank budget is critical for preserving task performance under a fixed compression constraint. We also compare different objectives for gradient-based whitening. Both LM loss and prefix KD underperform the proposed on-policy generation-aware distillation signal, confirming that gradients collected from the student’s own generated trajectories provide more effective compression guidance for generation tasks. Compared with teacher-forced or next-token prediction objectives, the on-policy signal reflects the distribution encountered by the compressed model during autoregressive decoding, where early errors can influence subsequent tokens.

%% file: tables/main_table.tex
\begin{table*}[t]
\centering
\small
\caption{Performance comparison of \sysname and baselines on both MCQ and generative benchmarks. The results are from Mistral-7B model with 20\% compression ratio. \sysname consistently outperforms SVD-based baselines across coding, math, MCQ benchmarks, and perplexity.}
\vspace{-.5em}
\resizebox{\textwidth}{!}{
\begin{tabular}{l ccc| c|ccccc|c}
\toprule

 \multirow{2}{*}{\textbf{Methods}}
& \multicolumn{3}{c|}{\textbf{Coding $\uparrow$}}
& \multicolumn{1}{c|}{\textbf{Math} $\uparrow$}
& \multicolumn{5}{c|}{\textbf{MCQ} $\uparrow$}
& \multirow{2}{*}{\textbf{PPL Avg} $\downarrow$} \\
\cmidrule(lr){2-6} \cmidrule(lr){6-6} \cmidrule(lr){6-10}
  
& \textbf{HumanEval} & \textbf{MBPP} & \textbf{Code Avg}
& \textbf{GSM8K}&\textbf{PIQA} &\textbf{MMLU} &\textbf{ARC-C}&\textbf{HS}&\textbf{Avg} \\
\midrule
 Baseline
 &29.3 &51.6 &40.4  &40.4 &82.1 &62.6 &61.3 &83.3 &72.3 &13.6  \\
\cmidrule(lr){1-11}
 
 ASVD~\citep{yuan2023asvd}
&0.0 &6.6  &3.3  &1.5  &67.4  &27.0 &28.6 &40.7&40.9&96.1 \\
 
 SVDLLM~\citep{wang2025svdllm}
&15.2 &28.6  &21.9  &14.9 &65.6 &39.8  &32.2 &44.3 &45.4 &50.4\\

 SVDLLM v2~\citep{wang2025svd2}
&18.9 &28.3  &23.6  &17.8  &68.5  &40.0  &35.6 &48.9 &48.2 &41.7\\

Dobi-SVD~\citep{wang2025dobi}
&15.2 &29.1  &22.1  &15.6  &69.1  &43.3  &33.6&51.7 &49.4 &39.0\\

AA-SVD~\cite{sinha2026aa}
&18.9 &31.7  &25.3  &15.5  &65.0  &40.9  &31.3 &44.4 &45.4 &63.2 \\
 \sysname (Ours)
&\textbf{20.7} & \textbf{34.1} & \textbf{27.4} & \textbf{18.5} & \textbf{69.9} & \textbf{47.9} & \textbf{36.1} & \textbf{53.0} & \textbf{51.7} & \textbf{37.8} \\
\bottomrule
\end{tabular}
}
\vspace{-1em}
\label{tab:main_table}
\end{table*}

%% file: tables/main_larger.tex
\begin{table*}[t]
\centering
\small
\caption{Performance of \sysname on different model series including Qwen3 and Llama3.3. \sysname consistently improves over SVDLLM on Qwen3 models from 8B to 32B and scales to Llama-3.3-70B-Instruct, improving both generation and MCQ performance while reducing perplexity.}
\vspace{-1em}
\resizebox{\textwidth}{!}{
\begin{tabular}{ll ccc| c|ccccc|c}
\toprule
\multirow{2}{*}{\textbf{Model}} 
& \multirow{2}{*}{\textbf{Methods}}
& \multicolumn{3}{c|}{\textbf{Coding} $\uparrow$}
& \multicolumn{1}{c|}{\textbf{Math} $\uparrow$}
& \multicolumn{5}{c|}{\textbf{MCQ} $\uparrow$}
& \multirow{2}{*}{\textbf{PPL Avg} $\downarrow$} \\
\cmidrule(lr){3-6} \cmidrule(lr){6-6} \cmidrule(lr){6-11}
&  
& \textbf{HumanEval} & \textbf{MBPP} & \textbf{Code Avg}
& \textbf{GSM8K}&\textbf{PIQA} &\textbf{MMLU} &\textbf{ARC-C}&\textbf{HS}&\textbf{Avg} \\

\midrule
\multirow{4}{*}{Qwen3-8B}
& Baseline &86.0 &70.9 &78.4  &88.7  &77.6 &74.9 &67.3 &77.6 &74.3 &13.5  \\
\cmidrule(lr){2-12}
& SVDLLM
&0.0 &0.0  &0.0  &1.4  &62.9  &34.8 &30.9 &45.6 &43.5 &46.0 \\

& \sysname (Ours)
& \textbf{9.1} & \textbf{23.8} & \textbf{16.5} & \textbf{8.4} & \textbf{65.4} & \textbf{45.4} & \textbf{33.3} & \textbf{47.1} & \textbf{47.8} & \textbf{31.2}\\

\midrule
\multirow{4}{*}{Qwen3-14B}
& Baseline &87.0 &89.0 &88.0  &91.2 &79.8 &78.8 &69.8 &79.9 &77.0 &12.0  \\
\cmidrule(lr){2-12}
& SVDLLM
&19.0 &18.9  &19.0  &11.6  &68.2  &50.3 &40.3 &51.5 &52.5 &30.0 \\

& \sysname (Ours)
&\textbf{44.4} &\textbf{43.9}  &\textbf{44.1}  &\textbf{30.3}  &\textbf{69.2}  &\textbf{53.5}  &\textbf{44.3} &\textbf{55.8} &\textbf{55.7} &\textbf{24.9}\\

\cmidrule(lr){1-12}
\multirow{4}{*}{Qwen3-32B}
& Baseline &91.5 &87.0 &89.2 &89.0  &79.1 &81.0 &71.6 &71.0 &75.6 &10.5  \\
\cmidrule(lr){2-12}
& SVDLLM
&36.6 &32.8  &34.7  &31.2  &72.0  &63.7 &50.3 &\textbf{65.4} &62.8 &22.9 \\
& \sysname (Ours)
&\textbf{62.2} &\textbf{65.9}  &\textbf{64.0}  &\textbf{63.1}  &\textbf{72.9}  &\textbf{64.2}  &\textbf{51.3} &64.2 &\textbf{63.1} &\textbf{18.3}\\

\cmidrule(lr){1-12}
\multirow{4}{*}{Llama-3.3-70B-Ins}
& Baseline &83.5 &86.8 &85.1  &94.2  &84.7 &82.4 &72.7 &86.4 &81.5 &7.2  \\
\cmidrule(lr){2-12}
& SVDLLM
&42.7 &56.9  &49.8  &55.1  &74.4  &58.7 &45.3 &64.7 &60.7 &22.9 \\
& \sysname (Ours)
& \textbf{68.4} & \textbf{51.8} & \textbf{60.1} & \textbf{63.0} & \textbf{76.7} & \textbf{63.5} & \textbf{52.9} & \textbf{70.6} & \textbf{65.9} & \textbf{18.5} \\
\bottomrule
\end{tabular}
}
\vspace{-1em}
\label{tab:large_scale}
\end{table*}

%% file: tables/table_ratio.tex
\begin{table*}[t]
\centering
\small
\caption{Performance on different compression ratios. \sysname consistently outperforms SVD-LLM at both 20\% and 40\% compression, showing stronger performance across coding, math, MCQ benchmarks, and perplexity.}
\vspace{-1em}
\resizebox{\textwidth}{!}{
\begin{tabular}{ll ccc| c|ccccc|c}
\toprule
\multirow{2}{*}{\textbf{Ratio}} 
& \multirow{2}{*}{\textbf{Methods}}
& \multicolumn{3}{c|}{\textbf{Coding} $\uparrow$}
& \multicolumn{1}{c|}{\textbf{Math} $\uparrow$}
& \multicolumn{5}{c|}{\textbf{MCQ} $\uparrow$}
& \multirow{2}{*}{\textbf{PPL Avg} $\downarrow$} \\
\cmidrule(lr){3-6} \cmidrule(lr){6-6} \cmidrule(lr){6-11}
&  
& \textbf{HumanEval} & \textbf{MBPP} & \textbf{Code Avg}
& \textbf{GSM8K}&\textbf{PIQA} &\textbf{MMLU} &\textbf{ARC-C}&\textbf{HS}&\textbf{Avg} \\

\midrule
\multirow{5}{*}{20\%}
&Baseline
 &29.3 &51.6 &40.4  &40.4 &82.1 &62.6 &61.3 &83.3 &72.3 &13.6  \\
\cmidrule(lr){2-12}
&  SVDLLM
&15.2 &28.6  &21.9  &14.9 &65.6 &39.8  &32.2 &44.3 &45.4 &50.4\\

& \sysname (Ours)
& \textbf{20.7} & \textbf{34.1} & \textbf{27.4} & \textbf{18.5} & \textbf{69.9} & \textbf{47.9} & \textbf{36.1} & \textbf{53.0} & \textbf{51.7} & \textbf{37.8}\\

\cmidrule(lr){2-12}
\multirow{2}{*}{40\%}& SVDLLM
&8.5 &11.1  &9.8  &5.3  &56.3  &27.5 &24.3 &31.8 &34.9 &137.4 \\

& \sysname (Ours)
& \textbf{10.2} & \textbf{16.1} & \textbf{13.1} & \textbf{6.2} & \textbf{58.4} & \textbf{28.9} & \textbf{25.9} & \textbf{34.9} & \textbf{37.0} & \textbf{97.5}\\
\bottomrule
\end{tabular}
}
\vspace{-2em}
\label{tab:compression_ratio}
\end{table*}

%% file: Sec/fig_sec/mistral_rank_allocation.tex
\begin{figure*}[t]
\begin{center}
\subfloat[Down\_Proj]{\includegraphics[width=0.33\textwidth]{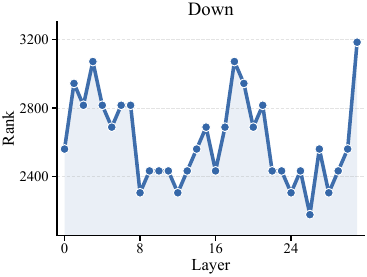}}
\subfloat[K\_Proj]{\includegraphics[width=0.33\textwidth]{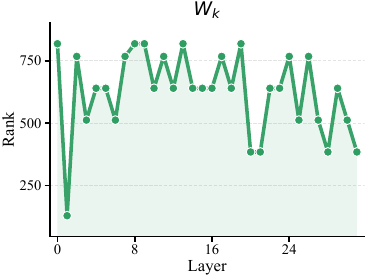}}
\subfloat[O\_Proj]{\includegraphics[width=0.33\textwidth]{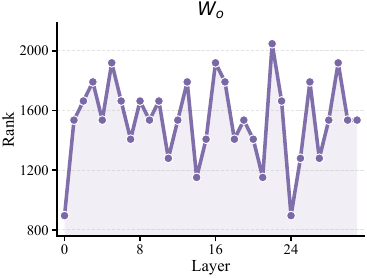}}\\ 
\end{center}
\vspace{-1.5em}
\caption{Layer-wise Rank Allocation.
The allocated ranks vary substantially across layers and weight matrices, indicating that uniform rank assignment is suboptimal for low-rank LLM compression.}
\label{Fig_exp:rank_alloc}
\vspace{-1em}

\end{figure*}

%% file: tables/speedup.tex
\begin{table*}[t]
\begin{minipage}{0.47\textwidth}
    \centering
    \caption{Throughput, memory usage, and compression runtime on a single H100 GPU. \sysname improves inference throughput and reduces GPU memory usage compared with the dense model, while requiring substantially less compression time than Dobi-SVD.}
    \vspace{-.5em}
    \resizebox{1\textwidth}{!}{
    \begin{tabular}{l|c|c|c}
    \toprule
    \textbf{Model}&\textbf{Throughput (Tokens/s)}&\textbf{Memory (GB)}  &\textbf{Runtime (h)}   \\ 
    \midrule
    Dense & 22931	&13.6  &-\\
    Dobi-SVD &-	&-  &112.4\\
    \sysname	&27720 (20.8\% $\uparrow$) 	&8.3 (38.9\% $\downarrow$) &5.2\\
    
    \bottomrule
    \end{tabular}
    }
    \label{tab:speedup}
    \vspace{-1em}
\end{minipage}%
\hfill
\begin{minipage}{0.47\textwidth}
    \centering
    \caption{Ablation Study. We compare the results without each component and with different loss choice for gradient computation. \sysname outperforms on all settings, demonstrating the effectiveness of each component.}
    \vspace{-.5em}
    \resizebox{0.85\textwidth}{!}{
    \begin{tabular}{l|cccc}
    \toprule
    \textbf{Model} & \textbf{HumanEval} & \textbf{MBPP}   & \textbf{GSM8K}   \\ 
    \midrule
    \sysname &\textbf{20.7} &\textbf{34.1} &\textbf{18.5}  \\
    \sysname w/o GAWW &19.1 &32.0 &18.1  \\
    \sysname w/o ERA &16.7 &30.9 &18.3  \\
    \cmidrule{1-4}
    LM Loss &18.9 &31.6 & 17.3  \\
    Prefix KD  &18.3  &32.2  &17.5   \\
    \bottomrule
    \end{tabular}
    }
    \label{tab:ablation}
    
\end{minipage}
\vspace{-1em}
\end{table*}

%% file: Sec/conclusion.tex
\section{Conclusion}

In this paper, we propose GA-SVD, a generation-aware low-rank compression framework for LLMs. GA-SVD introduces Generation-Aware Weight Whitening, which uses on-policy distillation gradients to align the compression geometry with generative behavior, and an evolutionary global rank allocation strategy to search for effective layer-wise ranks under a fixed compression budget. Experiments across Mistral, Qwen3, and Llama-3.3 models show that GA-SVD consistently outperforms existing SVD-based baselines on both generative and multiple-choice benchmarks. The method also reduces perplexity, improves inference efficiency, and scales to large models up to 70B parameters. These results demonstrate that generation-aware low-rank compression is an effective and practical direction for preserving LLM capabilities under substantial compression.

%% file: Sec/appendix.tex
\newpage
\appendix
\section{Appendix}

\subsection{Derivation of Generation-Aware Bi-Whitened Low-Rank Compression}
\label{app:gaww_derivation}

In this appendix, we provide the derivation underlying the main-text formulation of
Generation-Aware Weight Whitening (GAWW).
We first derive the bi-sided quadratic approximation of the layerwise loss increase, and then show how
the resulting weighted low-rank approximation problem reduces to ordinary truncated SVD in a transformed
space.

\paragraph{From output perturbation to local quadratic loss.}
Consider a linear layer $l$ with original weight $W_l^\star$ and compressed approximation $\widehat W_l$.
Define the compression error
\[
\Delta W_l = W_l^\star - \widehat W_l.
\]
For token position $t$, let $x_{l,t}$ denote the layer input and
\[
h_{l,t} = W_l^\star x_{l,t}
\]
denote the corresponding layer output.
Under compression, the perturbed output becomes
\[
\widehat h_{l,t} = \widehat W_l x_{l,t},
\]
and therefore the output perturbation is
\begin{equation}
\Delta h_{l,t} = h_{l,t} - \widehat h_{l,t} = \Delta W_l x_{l,t}.
\label{eq:app_output_perturb}
\end{equation}

Let $\ell_t$ denote the token-level generation-aware objective evaluated at token position $t$.
A second-order Taylor expansion around the original layer output gives
\begin{equation}
\ell_t(h_{l,t} - \Delta h_{l,t})
\approx
\ell_t(h_{l,t})
-
g_{l,t}^{\top}\Delta h_{l,t}
+
\frac{1}{2}\Delta h_{l,t}^{\top} H_{l,t}\Delta h_{l,t},
\label{eq:app_taylor}
\end{equation}
where
\[
g_{l,t} = \frac{\partial \ell_t}{\partial h_{l,t}},
\qquad
H_{l,t} = \frac{\partial^2 \ell_t}{\partial h_{l,t}\partial h_{l,t}^{\top}}.
\]
Thus, the compression-induced increase in token-level loss is approximated by
\begin{equation}
\Delta \ell_t
\approx
-g_{l,t}^{\top}\Delta h_{l,t}
+
\frac{1}{2}\Delta h_{l,t}^{\top} H_{l,t}\Delta h_{l,t}.
\label{eq:app_delta_loss}
\end{equation}

Following a standard empirical Fisher / Gauss--Newton approximation, we neglect the first-order term and
replace the output-side curvature by the second moment of output gradients.
Averaging over generated-token positions yields the output-side sensitivity covariance
\begin{equation}
\Sigma_{s,l}^{(m)}
=
\frac{1}{N_l^{(m)}}
\sum_{t=1}^{N_l^{(m)}} g_{l,t}^{(m)} g_{l,t}^{(m)\top},
\label{eq:app_sigma_s}
\end{equation}
and the input covariance
\begin{equation}
\Sigma_{x,l}^{(m)}
=
\frac{1}{N_l^{(m)}}
\sum_{t=1}^{N_l^{(m)}} x_{l,t}x_{l,t}^{\top}.
\label{eq:app_sigma_x}
\end{equation}
Using Eq.~\eqref{eq:app_output_perturb}, the expected layerwise loss increase becomes
\begin{equation}
\Delta \mathcal{L}_l(\widehat W_l)
\approx
\frac{1}{2}
\mathbb{E}_t\!\left[
(\Delta W_l x_{l,t})^{\top}
\Sigma_{s,l}^{(m)}
(\Delta W_l x_{l,t})
\right].
\label{eq:app_loss_expectation}
\end{equation}

Now use the identity
\[
a^{\top}Ba = \mathrm{Tr}(Baa^{\top}),
\]
which gives
\begin{align}
(\Delta W_l x_{l,t})^{\top}\Sigma_{s,l}^{(m)}(\Delta W_l x_{l,t})
&=
\mathrm{Tr}\!\left(
\Sigma_{s,l}^{(m)}
\Delta W_l x_{l,t}x_{l,t}^{\top}\Delta W_l^{\top}
\right).
\end{align}
Taking expectation over tokens and substituting Eq.~\eqref{eq:app_sigma_x} yields
\begin{equation}
\Delta \mathcal{L}_l(\widehat W_l)
\approx
\frac{1}{2}
\mathrm{Tr}\!\left(
\Sigma_{s,l}^{(m)}
\Delta W_l
\Sigma_{x,l}^{(m)}
\Delta W_l^{\top}
\right),
\label{eq:app_main_quad}
\end{equation}
which is exactly the bi-sided quadratic objective used in the main text.

\paragraph{Weighted low-rank approximation problem.}
Substituting $\Delta W_l = W_l^\star - \widehat W_l$ into Eq.~\eqref{eq:app_main_quad}, the layerwise
compression problem can be written as
\begin{equation}
\widehat W_l
=
\arg\min_{\mathrm{rank}(A)\le r_l}
\mathrm{Tr}\!\left(
\Sigma_{s,l}
\left(W_l^\star-A\right)
\Sigma_{x,l}
\left(W_l^\star-A\right)^{\top}
\right).
\label{eq:app_weighted_problem}
\end{equation}

To ensure numerical stability, we regularize the two covariances:
\begin{equation}
\overline{\Sigma}_{x,l} = \Sigma_{x,l} + \epsilon_x I,
\qquad
\overline{\Sigma}_{s,l} = \Sigma_{s,l} + \epsilon_s I.
\label{eq:app_reg_cov}
\end{equation}
Let
\begin{equation}
\overline{\Sigma}_{x,l} = R_{x,l}R_{x,l}^{\top},
\qquad
\overline{\Sigma}_{s,l} = R_{s,l}R_{s,l}^{\top}
\label{eq:app_factors}
\end{equation}
be square-root factorizations, obtained by Cholesky decomposition in practice.

Substituting Eq.~\eqref{eq:app_factors} into Eq.~\eqref{eq:app_weighted_problem} gives
\begin{align}
&\mathrm{Tr}\!\left(
R_{s,l}R_{s,l}^{\top}
(W_l^\star-A)
R_{x,l}R_{x,l}^{\top}
(W_l^\star-A)^{\top}
\right).
\label{eq:app_trace_expand}
\end{align}
Using cyclic invariance of the trace, this is equal to
\begin{align}
&\mathrm{Tr}\!\left(
R_{s,l}^{\top}(W_l^\star-A)R_{x,l}
\left(
R_{s,l}^{\top}(W_l^\star-A)R_{x,l}
\right)^{\top}
\right) \notag\\
&=
\left\|
R_{s,l}^{\top}(W_l^\star-A)R_{x,l}
\right\|_F^2.
\label{eq:app_frob_equiv}
\end{align}
Therefore, Eq.~\eqref{eq:app_weighted_problem} is equivalent to
\begin{equation}
\widehat W_l
=
\arg\min_{\mathrm{rank}(A)\le r_l}
\left\|
R_{s,l}^{\top}(W_l^\star-A)R_{x,l}
\right\|_F^2.
\label{eq:app_rank_frob}
\end{equation}

\paragraph{Transformed-space formulation.}
Define the transformed weight
\begin{equation}
\widetilde W_l
=
R_{s,l}^{\top} W_l^\star R_{x,l}.
\label{eq:app_transform}
\end{equation}
If we further define the transformed variable
\[
B = R_{s,l}^{\top} A R_{x,l},
\]
then Eq.~\eqref{eq:app_rank_frob} becomes
\begin{equation}
\min_{\mathrm{rank}(B)\le r_l}
\left\|
\widetilde W_l - B
\right\|_F^2.
\label{eq:app_B_problem}
\end{equation}
Since $R_{s,l}$ and $R_{x,l}$ are invertible, multiplication by them preserves rank, and thus
\[
\mathrm{rank}(A)\le r_l
\iff
\mathrm{rank}(B)\le r_l.
\]
Therefore, the original weighted compression problem is exactly reduced to ordinary rank-$r_l$
Frobenius-norm approximation of the transformed matrix $\widetilde W_l$.

\paragraph{Optimal truncated SVD solution.}
Let
\begin{equation}
\widetilde W_l
=
U_l \Lambda_l V_l^{\top}
\label{eq:app_svd}
\end{equation}
be the singular value decomposition of the transformed matrix.
By the Eckart--Young theorem, the optimal rank-$r_l$ approximation to Eq.~\eqref{eq:app_B_problem} is
obtained by truncating Eq.~\eqref{eq:app_svd} to its top-$r_l$ singular values:
\begin{equation}
\widetilde W_{l,r}
=
U_{l,r} \Lambda_{l,r} V_{l,r}^{\top}.
\label{eq:app_trunc}
\end{equation}
Mapping the solution back to the original parameterization gives
\begin{equation}
\widehat W_l
=
R_{s,l}^{-\top}
U_{l,r} \Lambda_{l,r} V_{l,r}^{\top}
R_{x,l}^{-1}.
\label{eq:app_reconstruct}
\end{equation}
Equivalently, the compressed layer can be expressed in factorized form
\begin{equation}
\widehat W_l = W_{u,l}W_{v,l},
\label{eq:app_factorized}
\end{equation}
where
\begin{equation}
W_{u,l}
=
R_{s,l}^{-\top}
U_{l,r} \Lambda_{l,r}^{1/2},
\qquad
W_{v,l}
=
\Lambda_{l,r}^{1/2} V_{l,r}^{\top} R_{x,l}^{-1}.
\label{eq:app_factor}
\end{equation}

\newpage  

\begin{table}[t]
\centering
\caption{Summary of notation used in GA-SVD.}
\label{tab:notation}
\small
\begin{tabular}{ll}
\toprule
\textbf{Notation} & \textbf{Description} \\
\midrule
$M_T$ & Full-precision teacher model. \\
$M_S$ & Compressed student model. \\
$M_S^{(0)}$ & Warm-start compressed student model. \\
$\mathcal{L}$ & Set of linear layers to compress. \\
$W_l^\star$ & Original full-precision weight matrix of layer $l$. \\
$W_{c,l}$ & Low-rank compressed approximation of $W_l^\star$. \\
$r_l$ & Retained rank of layer $l$. \\
$x_{l,t}$ & Input activation of layer $l$ at token position $t$. \\
$h_{l,t}$ & Output activation of layer $l$ at token position $t$. \\
$g_{l,t}$ & Gradient of the generation-aware objective with respect to $h_{l,t}$. \\
$\mathcal{L}_{\mathrm{gen}}$ & On-policy generation-aware distillation objective. \\
$\Sigma_{x,l}$ & Input covariance matrix of layer $l$. \\
$\Sigma_{s,l}$ & Generation-aware output sensitivity covariance matrix of layer $l$. \\
$R_{x,l}, R_{s,l}$ & Square-root factors of the regularized input and sensitivity covariance matrices. \\
$\widetilde{W}_l$ & Bi-transformed weight matrix. \\
$U_{l,r}, \Lambda_{l,r}, V_{l,r}$ & Truncated SVD factors of $\widetilde{W}_l$ with rank $r_l$. \\
$W_{u,l}, W_{v,l}$ & Factorized compressed weights with $W_{c,l}=W_{u,l}W_{v,l}$. \\
\midrule
$\mathcal{W}$ & Set of compressible weight matrices for rank allocation. \\
$K_m$ & Retained singular-index set of weight matrix $m$. \\
$B$ & Global compression budget. \\
$\rho$ & Target compression ratio. \\
$F(\cdot)$ & Fitness function used in evolutionary rank allocation. \\
$O$ & Genome encoding a candidate rank allocation. \\
\bottomrule
\end{tabular}
\end{table}

\subsection{Limitation}

Although \sysname achieves strong performance on both MCQ and generative tasks, several limitations remain. First, the actual inference speedup is lower than the nominal compression ratio. This is an inherent limitation of SVD-based compression, as each dense matrix is replaced by the product of two smaller matrices, introducing additional computation and memory access overhead. This limitation may be mitigated through system-level co-design and optimized kernels. Second, a noticeable generation performance gap still remains, particularly when compressing smaller models or under aggressive compression ratios. We leave further improving generation preservation and inference efficiency to future work.

\subsection{More Experimental Results}
\label{app:more_exp_res}

\paragraph{Results on CNN/DailyMail.} To complement the structured generation benchmarks, we evaluate open-ended summarization using Mistral-7B on 200 samples from CNN/DailyMail. As shown in Table~\ref{tab:summarization}, GA-SVD improves ROUGE-1 from 22.62 to 24.98 and ROUGE-2 from 9.64 to 11.50 over SVD-LLM, narrowing the gap to the dense model, which achieves 30.07 and 13.60, respectively. These results suggest that the benefits of generation-aware compression extend to summarization, helping preserve the content captured by ROUGE beyond tasks evaluated through code execution or exact-answer matching.

\begin{table}[t]
    \centering
    \caption{Performance on abstractive summarization.}
    \label{tab:summarization}
    \small
    \begin{tabular}{lcc}
        \toprule
        Method & ROUGE-1 $\uparrow$ & ROUGE-2 $\uparrow$ \\
        \midrule
        Dense   & 30.07 & 13.60 \\
        SVD-LLM & 22.62 & 9.64 \\
        GA-SVD  & \textbf{24.98} & \textbf{11.50} \\
        \bottomrule
    \end{tabular}
\end{table}

\paragraph{Results on more benchmarks}
\begin{table*}[t]
\centering
\small
\resizebox{0.8\textwidth}{!}{
\begin{tabular}{l ccc| cccccc}
\toprule

 \multirow{2}{*}{\textbf{Methods}}
& \multicolumn{3}{c|}{\textbf{PPL}$\downarrow$}
& \multicolumn{6}{c}{\textbf{MCQ} $\uparrow$} \\
\cmidrule(lr){2-4} \cmidrule(lr){5-10}
  
& \textbf{Wiki} & \textbf{C4} & \textbf{PTB}  &\textbf{ARC-E} &\textbf{ARC-C} &\textbf{PIQA} &\textbf{HS} &\textbf{MMLU} &\textbf{Avg} \\
\midrule
 Baseline
 &5.2 &8.1 &27.7   &79.5 &61.3 &28.1 &83.3 &62.6 &63.0 \\
\cmidrule(lr){1-10}
 
 ASVD
&20.5&30.8  &237.1    &53.4  &28.6 &67.4 &40.7 &27 &43.4 \\
 
 SVD-LLM
&12.4 &24.6  &114.2   &55.4 &32.2  &65.6 &44.3 &39.8&47.5\\

 SVD-LLM v2
&\textbf{10.6} &22.8  &91.7    &63.5  &35.6  &68.5 &48.9 &40.0&51.3\\

Dobi-SVD
&14.9 &21.8 &\textbf{80.1}   &\textbf{64.6}  &33.6  &69.1 &51.7 &43.3 &49.4 \\

AA-SVD
&24.2 &36.5  &128.9    &50.8 &31.3  &65.0 &44.4 &40.9 &46.5 \\
 \sysname (Ours)
&11.2 &\textbf{20.8} &81.6   &61.8 &\textbf{36.1}  &\textbf{69.9} &\textbf{53.0} &\textbf{47.9}&\textbf{53.7}\\
\bottomrule
\end{tabular}
}
\caption{Detailed perplexity and MCQ benchmark results on Mistral-7B with 20\% compression.
We report perplexity on WikiText, C4, and PTB, together with additional MCQ evaluations on ARC-E, ARC-C, PIQA, HellaSwag, and MMLU. GA-SVD achieves the best MCQ average among compressed models while maintaining competitive perplexity, demonstrating a strong overall trade-off between language modeling quality and downstream task performance.}
\label{tab:more_main_table}
\end{table*}

Table ~\ref{tab:more_main_table} provides a more detailed comparison of perplexity and MCQ performance on Mistral-7B under 20\% compression. GA-SVD achieves the best MCQ average among all compressed models, showing that the proposed method better preserves discriminative reasoning ability after low-rank compression. In particular, GA-SVD obtains the best results on ARC-C, PIQA, HellaSwag, and MMLU, and remains competitive on ARC-E. These consistent gains across diverse MCQ benchmarks indicate that generation-aware whitening does not only benefit open-ended generation tasks, but also improves general downstream reasoning performance. For perplexity, GA-SVD also shows a favorable trade-off. It achieves the best C4 perplexity among compressed models and remains close to the best compressed result on WikiText and PTB. Compared with baselines that suffer severe perplexity degradation, GA-SVD better preserves the language modeling distribution while improving downstream accuracy. Overall, these results further confirm that GA-SVD provides a stronger compression-quality trade-off than prior SVD-based methods, maintaining both language modeling quality and benchmark performance under the same compression budget.

\paragraph{Results with Different Student Warmup Ratio.}
We evaluate the sensitivity to the warmup ratio by varying it from 5\% to 20\%. As shown in Table~\ref{tab:warmup_ratio}, final performance remains stable across this range: the coding average ranges from 27.4 to 28.1, GSM8K accuracy from 18.2 to 19.2, and the MCQ average from 51.7 to 51.8. Increasing the warmup ratio yields no consistent improvement, suggesting that a short warmup is sufficient for effective estimation and that the method is insensitive to the warmup ratio within the evaluated range.

\begin{table}[t]
    \centering
    \caption{Performance with different warmup ratios.}
    \label{tab:warmup_ratio}
    \small
    \setlength{\tabcolsep}{4pt}
    \begin{tabular}{lccccccccc}
        \toprule
        Warmup Ratio & HumanEval & MBPP & Avg & GSM8K
        & PIQA & MMLU & ARC-C & HS & Avg \\
        \midrule
        5\%  & 21.5 & 34.8 & 28.1 & 19.2
             & 68.6 & 48.2 & 36.5 & 53.6 & 51.7 \\
        10\% & 21.0 & 34.1 & 27.5 & 18.2
             & 70.2 & 47.5 & 37.1 & 52.4 & 51.8 \\
        20\% & 20.7 & 34.1 & 27.4 & 18.5
             & 69.9 & 47.9 & 36.1 & 53.0 & 51.7 \\
        \bottomrule
    \end{tabular}
\end{table}

\paragraph{Combination with Quantization.} 
We further evaluate the compatibility of GA-SVD with quantization by applying 4-bit weight-only GPTQ to the compressed model. As shown in Table~\ref{tab:compression_methods}, quantization introduces modest performance degradation: the coding average decreases from 27.4 to 25.8, GSM8K accuracy from 18.5 to 17.8, and the MCQ average from 51.7 to 50.5, corresponding to drops of 1.6, 0.7, and 1.2 percentage points, respectively. The quantized model also retains higher coding, GSM8K, and MCQ average scores than the evaluated Wanda and SparseGPT baselines. These results suggest that GA-SVD can be effectively combined with low-bit quantization, enabling further compression of the model weights while largely preserving downstream performance.
\begin{table}[t]
    \centering
    \caption{Performance comparison of different compression methods.}
    \label{tab:compression_methods}
    \small
    \setlength{\tabcolsep}{4pt}
    \begin{tabular}{lccccccccc}
        \toprule
        Methods & HumanEval & MBPP & Avg & GSM8K
        & PIQA & MMLU & ARC-C & HS & Avg \\
        \midrule
        Wanda
        & 4.3 & 4.8 & 4.5 & 1.4
        & 70.1 & 24.8 & 27.1 & 37.6 & 39.9 \\
        SparseGPT
        & 0 & 0 & 0 & 3.2
        & 76.9 & 38.4 & 32.5 & 43.2 & 47.7 \\
        GA-SVD
        & 20.7 & 34.1 & 27.4 & 18.5
        & 69.9 & 47.9 & 36.1 & 53.0 & 51.7 \\
        GA-SVD + GPTQ 4bit
        & 19.3 & 32.4 & 25.8 & 17.8
        & 68.2 & 46.2 & 35.7 & 52.1 & 50.5 \\
        \bottomrule
    \end{tabular}
\end{table}

\paragraph{Ablation on Calibration Samples and Rollout Length.}
We examine the sensitivity of GA-SVD to the number of calibration samples and rollout length in Table~\ref{tab:calibration_rollout}. Reducing the calibration set from 256 to 128 samples while keeping the rollout length at 512 leads to modest decreases of 0.9, 0.6, and 0.7 percentage points in the coding average, GSM8K accuracy, and MCQ average, respectively. With 256 calibration samples, shortening the rollout length from 512 to 256 yields comparable performance, with all three metrics changing by at most 0.3 percentage points. These results suggest that GA-SVD remains effective with a reduced calibration budget, and that shorter rollouts offer a practical way to reduce calibration cost with little impact on downstream performance.

\begin{table}[t]
    \centering
    \caption{Sensitivity to the number of calibration samples and rollout length.}
    \label{tab:calibration_rollout}
    \small
    \setlength{\tabcolsep}{4pt}
    \begin{tabular}{ccccccccccc}
        \toprule
        \multicolumn{2}{c}{Calibration Settings}
        & HumanEval & MBPP & Avg & GSM8K
        & PIQA & MMLU & ARC-C & HS & Avg \\
        \cmidrule(lr){1-2}
        Samples & Rollout Length
        & & & & & & & & & \\
        \midrule
        256 & 512
        & 20.7 & 34.1 & 27.4 & 18.5
        & 69.9 & 47.9 & 36.1 & 53.0 & 51.7 \\
        128 & 512
        & 19.3 & 33.7 & 26.5 & 17.9
        & 69.1 & 46.3 & 36.2 & 52.6 & 51.0 \\
        256 & 256
        & 19.7 & 34.5 & 27.1 & 18.6
        & 70.2 & 47.2 & 37.2 & 52.8 & 51.8 \\
        \bottomrule
    \end{tabular}
\end{table}

\paragraph{The Results on Qwen3-8B.}

\begin{table}[t]
    \centering
    \caption{Performance of Qwen3-8B at different compression ratios.}
    \label{tab:qwen3_8b_compression}
    \small
    \setlength{\tabcolsep}{4pt}
    \begin{tabular}{llccccccccc}
        \toprule
        Method & Compression & HumanEval & MBPP & Avg
        & GSM8K & PIQA & MMLU & ARC-C & HS & Avg \\
        \midrule
        SVD-LLM & 10\%
        & 21.1 & 20.5 & 20.8 & 20.3
        & 63.6 & 47.2 & 35.8 & 50.1 & 49.1 \\
        GA-SVD & 10\%
        & 31.0 & 39.6 & 35.3 & 28.2
        & 69.0 & 53.0 & 42.9 & 54.6 & 54.8 \\
        \midrule
        SVD-LLM & 20\%
        & 0.0 & 0.0 & 0.0 & 1.4
        & 62.9 & 34.8 & 30.9 & 45.6 & 43.5 \\
        GA-SVD & 20\%
        & 9.1 & 23.8 & 16.5 & 8.4
        & 65.4 & 45.4 & 33.3 & 47.1 & 47.8 \\
        \bottomrule
    \end{tabular}
\end{table}
Qwen3-8B exhibits pronounced sensitivity to low-rank compression, particularly on generation tasks. We provide results of different compression ratio on Qwen3-8B in Table~\ref{tab:qwen3_8b_compression}. At 20\% compression, the coding average drops from 78.4 to 0.0 with SVD-LLM and to 16.5 with GA-SVD, while GSM8K accuracy falls from 88.7 to 1.4 and 8.4, respectively. A possible explanation is that Qwen3-8B has less parameter redundancy than its larger counterparts, making the same relative rank reduction more disruptive to the capacity needed for autoregressive reasoning. The larger degradation on coding and math tasks than on MCQ benchmarks is consistent with this hypothesis, although it does not establish the underlying mechanism. Reducing compression to 10\% substantially alleviates the degradation: coding averages recover to 20.8 and 35.3, and GSM8K accuracies to 20.3 and 28.2, for SVD-LLM and GA-SVD, respectively. These results show that GA-SVD mitigates, but does not eliminate, the sensitivity of Qwen3-8B to compression, highlighting the need to adapt compression ratios to model capacity and task requirements.
\subsection{More Visualization of Rank Allocation}
\label{sec:more_vis}
\begin{figure*}[h]
\begin{center}
\subfloat[Down\_Proj]{\includegraphics[width=0.33\textwidth]{figures/rank_down_proj.pdf}}
\subfloat[Gate\_Proj]{\includegraphics[width=0.33\textwidth]{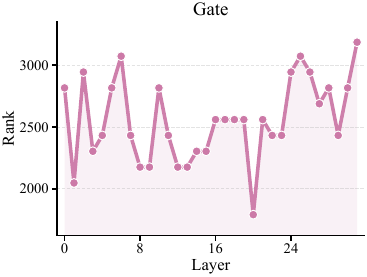}}
\subfloat[Up\_Proj]{\includegraphics[width=0.33\textwidth]{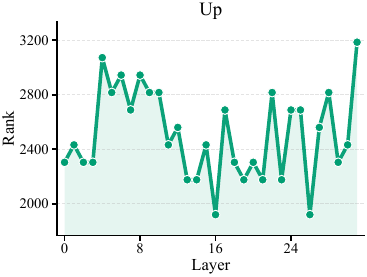}}\\

\subfloat[Q\_Proj]{\includegraphics[width=0.33\textwidth]{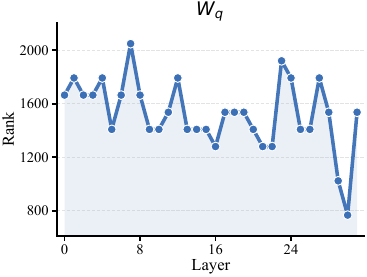}}
\subfloat[K\_Proj]{\includegraphics[width=0.33\textwidth]{figures/rank_k_proj.pdf}}
\subfloat[V\_Proj]{\includegraphics[width=0.33\textwidth]{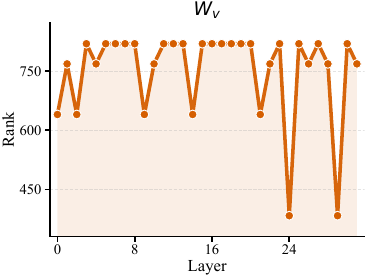}}
\end{center}

\vspace{-1em}
\caption{Layer-wise Rank Allocation of Mistral-7B.}
\label{Fig_exp:rank_alloc_mistral}

\end{figure*}

\begin{figure*}[h]
\begin{center}
\subfloat[Down\_Proj]{\includegraphics[width=0.33\textwidth]{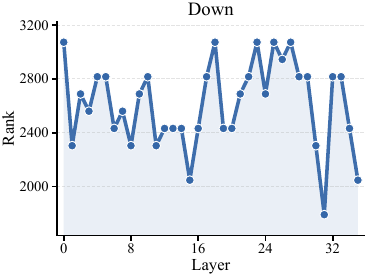}}
\subfloat[Gate\_Proj]{\includegraphics[width=0.33\textwidth]{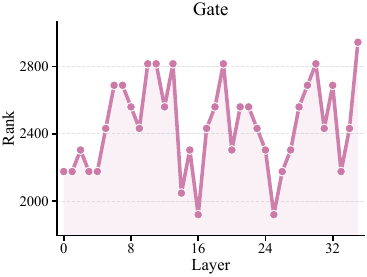}}
\subfloat[Up\_Proj]{\includegraphics[width=0.33\textwidth]{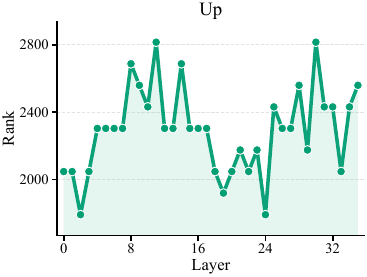}}\\

\subfloat[Q\_Proj]{\includegraphics[width=0.33\textwidth]{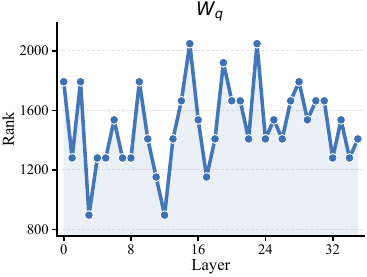}}
\subfloat[K\_Proj]{\includegraphics[width=0.33\textwidth]{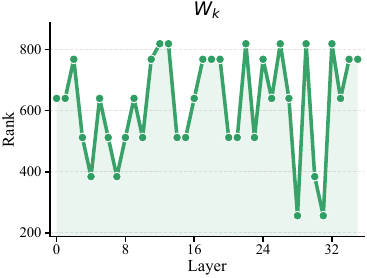}}
\subfloat[V\_Proj]{\includegraphics[width=0.33\textwidth]{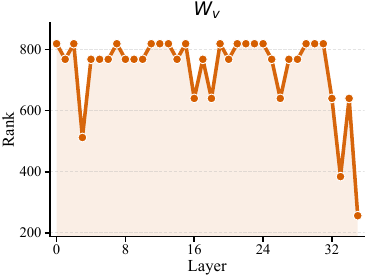}}
\end{center}

\vspace{-1em}
\caption{Layer-wise Rank Allocation of Qwen3-8B.}
\label{Fig_exp:rank_alloc_qwen3}

\end{figure*}

Figures ~\ref{Fig_exp:rank_alloc_mistral} and  ~\ref{Fig_exp:rank_alloc_qwen3} provide more detailed layer-wise rank allocation results for Mistral-7B and Qwen3. The learned ranks are highly non-uniform across both layers and weight types, further confirming that different parts of the transformer exhibit distinct sensitivity to low-rank compression. For example, MLP projections such as Down\_Proj, Gate\_Proj, and Up\_Proj generally receive much larger ranks than attention projections such as K\_Proj and V\_Proj, suggesting that MLP weights require more rank capacity to preserve model behavior under compression. The allocation patterns also differ across model families. Although both Mistral and Qwen3 show strong rank variation, the peaks and valleys occur at different layers and projection matrices, indicating that compression sensitivity is architecture-dependent. This supports the need for a model-specific global search strategy rather than using a fixed rank schedule shared across models. In addition, the sharp rank changes in some projections, such as Q\_Proj, K\_Proj, and V\_Proj, show that even weights within the same module can have substantially different redundancy. Overall, these results provide further evidence that GA-SVD’s evolutionary global rank allocation can effectively adapt the rank budget to both layer-wise and weight-wise heterogeneity.

